%% file: main_file.tex
\documentclass[sigconf]{acmart}

\copyrightyear{2026}
\acmYear{2026}
\setcopyright{cc}
\setcctype{by-nc-nd}
\acmConference[KDD '26]{Proceedings of the 32nd ACM SIGKDD Conference on Knowledge Discovery and Data Mining V.2}{August 09--13, 2026}{Jeju Island, Republic of Korea}
\acmBooktitle{Proceedings of the 32nd ACM SIGKDD Conference on Knowledge Discovery and Data Mining V.2 (KDD '26), August 09--13, 2026, Jeju Island, Republic of Korea}
\acmDOI{10.1145/3770855.3817715}
\acmISBN{979-8-4007-2259-2/2026/08}

\input{others/packages}
\input{others/configs}

\input{others/catppuccin_latte}

\begin{document}

\title[\model: Compressed and Composable KV Cache Reuse for Efficient LLM Inference]{\model: Compressed and Composable KV Cache Reuse \\ for Efficient LLM Inference}

\input{others/authors}

% CCS concepts



\input{text/0_abstract}

\maketitle

\newcommand\kddavailabilityurl{https://github.com/s7a9/C2KV}
\ifdefempty{\kddavailabilityurl}{}{
\begingroup\small\noindent\raggedright\textbf{Resource Availability:}\\
% please change the following context to include multiple artifacts if necessary, including data, models, code, etc.
The source code of this paper has been made publicly available at \url{\kddavailabilityurl}.
\endgroup
}

% The default list of authors is too long for headers.
\renewcommand{\shortauthors}{Chuheng Du et al.}

\input{text/1_intro_new}
\input{text/2_background}
\input{text/3_framework}
\input{text/4_experiments}

\input{text/5_related}
\input{text/6_conclusion}

\section*{Acknowledgements}
This work was sponsored in part by China NSF grant No. 62441236, 62472278, 62432007, 62332014, 62332013, and 62372296, and Shanghai QiYuan Innovation Foundation. 
This work was partially supported by Alibaba Group through the Alibaba Innovation Research Program and in part by SJTU Kunpeng \& Ascend Center of Excellence.
The opinions, findings, conclusions, and recommendations in this paper are those of the authors and do not necessarily reflect the views of the funding agencies or the government.

\newpage
\bibliographystyle{ACM-Reference-Format}
\balance
\bibliography{reference}

% \newpage
% \clearpage 
% \onecolumn 
\appendix
\input{appendix/appendix_outline}
\input{appendix/setup}
\input{appendix/ttft_measurement}
\input{appendix/full_result}
\input{appendix/limitations}

\end{document}

%% file: others/packages.tex
\usepackage{balance}
\usepackage{array}
\usepackage{mathrsfs}
\usepackage{amsmath,amsfonts,amsthm}
\usepackage{enumitem} 
\usepackage{graphicx}
\usepackage{extarrows}
\usepackage{subcaption}
\usepackage{verbatim}
\usepackage{makecell}
\usepackage{diagbox}
\usepackage{bm}
\usepackage{multirow}
\usepackage{float}
\usepackage{placeins}
\usepackage{xspace}
\usepackage[linesnumbered, ruled]{algorithm2e}
\usepackage{epstopdf}
\usepackage{setspace}
\usepackage[super]{nth}
\usepackage{fancyhdr}
\usepackage{slashbox}
\usepackage{lipsum,multicol}
\usepackage{url}
\usepackage[table,xcdraw]{xcolor}
\usepackage{booktabs}
\usepackage{diagbox}
\usepackage{caption}
\usepackage{ragged2e}
\usepackage{wrapfig}
\usepackage{tikz}
\usepackage{hyperref}
\usepackage{graphicx}
\usepackage[normalem]{ulem}
\usepackage{tabularx}
\usepackage{tikz}     % 用于绘制带圈数字

\usepackage{microtype}
\usepackage{graphicx}
\usepackage{subcaption}
\usepackage{booktabs} % for professional tables

\usepackage{hyperref}

\usepackage[capitalize,noabbrev]{cleveref}

\usepackage[textsize=tiny]{todonotes}

%% file: others/configs.tex
\theoremstyle{plain}

\theoremstyle{definition}

\theoremstyle{remark}

\def \eg {\emph{e.g.}, }

\def\BibTeX{{\rm B\kern-.05em{\sc i\kern-.025em b}\kern-.08em
    T\kern-.1667em\lower.7ex\hbox{E}\kern-.125emX}}

\SetCommentSty{mycommfont} 

\useunder{\uline}{\ul}{}

\newcommand{\model}{$\text{C}^2\text{KV}$\xspace} 
\newcommand{\method}{\model}
\newcommand{\methodprefix}{$\text{C}^2$}
\allowdisplaybreaks[4]

\newcommand{\mypara}[1]{\vspace{0.05cm}\noindent{\bf {#1}.}~}

%% file: others/catppuccin_latte.tex
\usepackage{xcolor}

\definecolor{Rosewater}{HTML}{dc8a78}
\definecolor{Flamingo}{HTML}{dd7878}
\definecolor{Pink}{HTML}{ea76cb}
\definecolor{Mauve}{HTML}{8839ef}
\definecolor{Red}{HTML}{d20f39}
\definecolor{Maroon}{HTML}{e64553}
\definecolor{Peach}{HTML}{fe640b}
\definecolor{Yellow}{HTML}{df8e1d}
\definecolor{Green}{HTML}{40a02b}
\definecolor{Teal}{HTML}{179299}
\definecolor{Sky}{HTML}{04a5e5}
\definecolor{Sapphire}{HTML}{209fb5}
\definecolor{Blue}{HTML}{1e66f5}
\definecolor{Lavender}{HTML}{7287fd}
\definecolor{CText}{HTML}{4c4f69}     % 文本颜色
\definecolor{Subtext1}{HTML}{5c5f77}
\definecolor{Subtext0}{HTML}{6c6f85}
\definecolor{Overlay2}{HTML}{7c7f93}
\definecolor{Overlay1}{HTML}{8c8fa1}
\definecolor{Overlay0}{HTML}{9ca0b0}
\definecolor{Surface2}{HTML}{acb0be}
\definecolor{Surface1}{HTML}{bcc0cc}
\definecolor{Surface0}{HTML}{ccd0da}
\definecolor{Base}{HTML}{eff1f5}      % 基础背景色
\definecolor{Mantle}{HTML}{e6e9ef}    % 次级背景
\definecolor{Crust}{HTML}{dce0e8}     % 最深背景

%% file: others/authors.tex
%\author{Shengzhong Liu}
%\affiliation{
% \institution{University of Illinois at Urbana-Champaign}
% \email{}{sl29@illinois.edu}}
% 
%\author{Tianshi Wang}
%\affiliation{
% \institution{University of Illinois at Urbana-Champaign}
% \email{}{tianshi3@illinois.edu}
%}
% 
%\author{Tarek Abdelzaher}
%\affiliation{
% \institution{University of Illinois at Urbana-Champaign}
% \email{}{zaher@illinois.edu}
%}

\author{Chuheng Du}
\orcid{0009-0005-3465-4023}
\affiliation{
\department{School of Computer Science}
\institution{Shanghai Jiao Tong University}
\city{Shanghai}
\country{China}
}
\email{dch7723@sjtu.edu.cn}

\author{Junyi Chen}
\affiliation{
\department{School of Computer Science}
\institution{Shanghai Jiao Tong University}
\city{Shanghai}
\country{China}
}
\email{junyi.chen@sjtu.edu.cn}

\author{Hanlin Tang}
\affiliation{
\institution{Alibaba Group}
\city{Hangzhou}
\country{China}
}
\email{tanghl1994@gmail.com}

\author{Kan Liu}
\affiliation{
\institution{Alibaba Group}
\city{Hangzhou}
\country{China}
}
\email{liukan.lk@alibaba-inc.com}

\author{Tao Lan}
\affiliation{
\institution{Alibaba Group}
\city{Hang Zhou}
\state{Zhejiang Province}
\country{China}
}
\email{tao.lant@alibaba-inc.com}

\author{Lin Qu}
\affiliation{
\institution{Alibaba Group}
\city{Hangzhou}
\country{China}
}
\email{xide.ql@taobao.com}

\author{Chaoyue Niu}
\affiliation{
\department{School of Computer Science}
\institution{Shanghai Jiao Tong University}
\city{Shanghai}
\country{China}
}
\email{rvince@sjtu.edu.cn}

\author{Shengzhong Liu}
\authornote{Shengzhong Liu is the corresponding author.}
\affiliation{
\department{School of Computer Science}
\institution{Shanghai Jiao Tong University}
\city{Shanghai}
\country{China}
}
\email{shengzhong@sjtu.edu.cn}

\author{Guihai Chen}
\affiliation{
\department{School of Computer Science}
\institution{Shanghai Jiao Tong University}
\city{Shanghai}
\country{China}
}
\email{gchen@cs.sjtu.edu.cn}

\author{Fan Wu}
\affiliation{
\department{School of Computer Science}
\institution{Shanghai Jiao Tong University}
\city{Shanghai}
\country{China}
}
\email{fwu@cs.sjtu.edu.cn}

%% file: text/0_abstract.tex
\begin{abstract}
Long-context inference is central to modern large language model (LLM) applications such as retrieval-augmented generation. To mitigate the growing inference cost, recent work has explored non-prefix key-value (KV) cache reuse to reduce redundant prefill computation. However, existing reuse methods primarily focus on computation savings and overlook a critical bottleneck in long-context LLM serving: the cost of storing and accessing large KV caches. While KV compression appears to be a natural complement, naively combining compression with non-prefix KV reuse often leads to severe accuracy degradation. %, as standard compressed KV representations are not designed to be composable across contexts. 
In this work, we propose \method, a unified framework for non-prefix KV reuse that jointly optimizes KV cache compression and concatenation. % Instead of compressing context-dependent KV states produced by standard attention,
\method learns a composable and compressed KV cache manifold that is explicitly designed to be position-agnostic. Our approach introduces a lightweight sidecar Extractor with learnable compression tokens and a structured attention flow, enabling modular KV representations that can be flexibly reused and concatenated without modifying the frozen base model. We further employ a compression-concatenation co-training strategy to align extraction-time representations with their downstream reuse behavior. Extensive experiments across multiple long-context benchmarks and model families demonstrate that \method significantly reduces KV cache storage and transfer costs, achieving up to 17$\times$ inference speedup under long contexts, while preserving generation quality.
% Large language models (LLMs)increasingly relies on long-context in-context learning (ICL) in real-world systems, such as retrieval-augmented generation (RAG), agentic systems, and few-shot learning. 
% In these workloads, long documents often repeat across requests but appear at different positions in the input. This leads to substantial redundant computation, large KV cache footprints, and high memory-bandwidth pressure during inference.
% Existing KV cache reuse techniques are largely limited to prefix matching and fail to exploit most document recurrence in practice.
% This paper studies non-prefix KV cache reuse for long-context inference. We analyze prior approaches and show that they either rely on approximate KV cache composition, which introduces irreducible accuracy loss, or require modifying the base model architecture, which degrades generalization and portability.
% We propose \model, a new non-prefix KV cache reuse framework that keeps the base model frozen and trains only a lightweight extraction module. 
% \model encodes documents into compressed, directly composable KV caches that can be reused at arbitrary positions without blending or recomputation. As a result, \model jointly reduces prefill computation, KV cache storage, and attention-time memory bandwidth while preserving generation quality.
% \red{Experiments on multi-document long-context benchmarks show that our method is effective under realistic workloads.}
% %\shengzhong{Add quantified experiment results summary. Address that our approach is generally compatible with multiple SOTA LLM models.}
\end{abstract}

%% file: text/1_intro_new.tex
\section{Introduction}
Large language models (LLMs) have been widely adopted across a broad range of applications, including question answering~\cite{chen-etal-2024-llm,yasunaga-etal-2021-qa, chen2025pre}, personal assistants~\cite{taylor2022galacticalargelanguagemodel, chen2025tokenflow} and knowledge-intensive reasoning systems~\cite{10.5555/3666122.3668302}.
In practical deployments, these applications are increasingly powered by \textit{in-context learning} (ICL) paradigms such as retrieval-augmented generation (RAG)~\cite{10.1145/3711896.3737012,NEURIPS2020_6b493230,10.1145/3637528.3671470,guo-etal-2025-lightrag}, agent systems with long-term memory~\cite{tan-etal-2025-prospect}, and few-shot learning~\cite{NEURIPS2020_1457c0d6}.
A common characteristic of these workloads is the continuous growth of input context length~\cite{lin2024infinite,ding2024longrope,xiao2024duoattention}, as each request injects multiple long documents into the model input.

% \junyi{Here should be a figure to show the target scenery and usage of our framework.}

\begin{figure}[t!]
    \centering
    \includegraphics[width=0.95\linewidth]{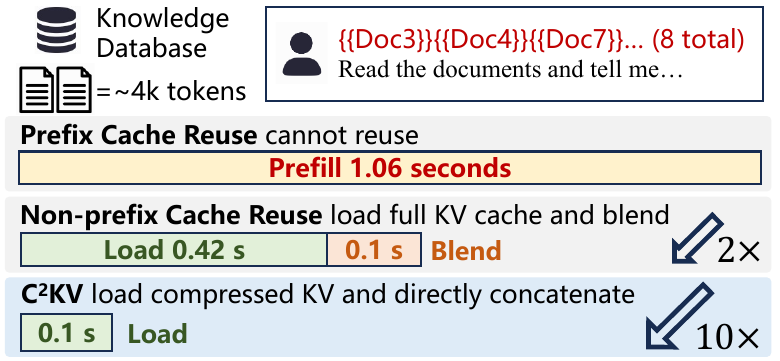}
    \caption{Breakdown of Time-to-First-Token (TTFT) in long-context in-context learning (ICL).}
    \label{fig:ttft_breakdown}
\end{figure}

As context lengths increase, LLM inference becomes prohibitively expensive. Given an input sequence of length $L$, inference incurs a compute-intensive prefill stage with $O(n^2)$ complexity to construct key-value (KV) caches, followed by autoregressive decoding where each generation step attends to all cached tokens, resulting in $O(L)$ memory storage and bandwidth costs. Consequently, long-context inference simultaneously stresses GPU compute, memory capacity, and memory bandwidth, motivating the need for effective KV cache reuse mechanisms in LLM serving systems.

KV cache reuse has therefore attracted significant attention.
Existing approaches can be broadly categorized into three classes.
1) \textit{Prefix-based reuse} exploits identical prefixes across requests and enables exact KV reuse under standard causal attention; however, its applicability is severely limited in ICL workloads where repeated documents rarely appear at the same position.
2) To go beyond prefix matching, \textit{training-free non-prefix} methods attempt to concatenate independently computed KV caches with limited recomputation to mitigate context mismatch, reducing prefill computation while preserving the original model.
3) In contrast, \textit{training-based} approaches modify the attention structure or introduce auxiliary tokens during pretraining to enable more flexible reuse or compression, at the cost of altering the base model or introducing information bottlenecks.

%However, existing KV cache reuse techniques primarily focus on reducing computation, while overlooking a critical challenge in real-world LLM serving systems: \textbf{the dominant bottleneck has shifted to KV cache storage and memory bandwidth.}
%In modern deployments, KV caches are often stored in lower-level memory (\eg host memory or secondary storage) and dynamically transferred to high-bandwidth GPU memory during inference. As context lengths grow, the cost of moving large KV caches across memory hierarchies and the increasing price of memory capacity itself becomes a major limiting factor, even when prefill computation is reduced.

%Unfortunately, existing non-prefix KV reuse methods do not directly address this storage and bandwidth bottleneck. While they enable reuse with limited recomputation, they still require storing and accessing large, uncompressed KV caches. 
%A natural intuition is to perform KV compression during the reuse procedure.
%However, naively applying compression to reused KV caches leads to significant accuracy degradation, as compressed representations are not designed to be composable across different contexts.

While existing KV cache reuse techniques are effective at reducing recomputation, they largely overlook a growing bottleneck in long-context LLM inference: the cost of storing and accessing KV caches.
In practical deployments, KV caches are often stored outside high-bandwidth accelerator memory and transferred on demand, making KV cache size and memory bandwidth critical factors as context length increases.
As a result, previous works overlook a critical challenge in real-world LLM serving systems: \textbf{the dominant bottleneck has shifted to KV cache storage and memory bandwidth.}

% This naturally motivates incorporating KV compression~\cite{hooper2024kvquant,zhang2024kv,tan2024alignedkv,zhang2023ho,10.5555/3692070.3694498} into non-prefix KV reuse.
% However, this exposes two distinct issues. First, compressed KV representations are context-dependent and non-composable, leading to inevitable accuracy degradation when reused or concatenated across different contexts.
% Second, there is an architectural mismatch. Long-context serving requires compression along the sequence dimension, whereas token-eviction methods perform layer- and head-dependent token selection. This design is incompatible with reuse as evicted tokens cannot be consistently aligned across heads and layers once the context changes. Under the requirement of architectural compatibility, we use SnapKV~\cite{10.5555/3737916.3738638} as a representative comparison, for subsequent token-eviction methods largely focus on refining per-layer or per-head budget allocation rather than addressing this core incompatibility.
This naturally motivates incorporating KV compression~\cite{hooper2024kvquant,zhang2024kv,tan2024alignedkv,zhang2023ho,10.5555/3692070.3694498} into non-prefix KV reuse.
However, this exposes two fundamental challenges.
First, most existing compression schemes produce KV representations that are tightly coupled to their original context and thus non-composable, leading to significant accuracy degradation when reused or concatenated across different contexts.
Second, there exists an architectural mismatch: long-context serving requires sequence-level compression, whereas many token-eviction methods rely on layer- or head-dependent selection, preventing consistent alignment once contexts are recomposed. As a result, even under the most reuse-compatible setting, \textbf{naive compression remains fundamentally incompatible with non-prefix KV reuse}, motivating a composability-aware design.
% However, existing KV compression methods are designed for fixed-context attention and produce representations that are tightly coupled to token order and attention structure.
% When such compressed KV states are reused or concatenated across different contexts, these assumptions no longer hold, leading to substantial accuracy degradation.
% This reveals a fundamental mismatch: \textbf{non-prefix KV reuse requires KV representations to be modular and composable, whereas existing compression is context-dependent and non-composable}.

To resolve this tension, we propose \method, a unified framework for non-prefix KV reuse that jointly optimizes the KV extraction and inference-time concatenation, as shown in Figure~\ref{fig:ttft_breakdown}. Unlike prior methods that directly compress or cache the original KV states, which are inherently context-dependent, \textbf{\method constructs a Composable and Compressed KV cache manifold.} This manifold is specifically designed to be \textit{position-agnostic}, allowing KV segments of various documents to be flexibly ``\textit{plugged and played}'' at arbitrary positions within a long-context prompt. 

The core of \method is its \methodprefix Extractor, a lightweight sidecar module that operates independently of the frozen base LLM. Unlike traditional fine-tuning, this ``attached'' module performs extraction without perturbing the model's original parametric knowledge. 
To achieve efficient reuse, the \methodprefix Extractor injects learnable \methodprefix Tokens as compressed memory slots, whose interactions are governed by a novel \textit{Structured Attention Flow}. This mechanism serves a dual purpose: first, it acts as a high-fidelity semantic aggregator that maximizes information retention via localized grouping; second, it ensures the resulting KV representations are modular and position-independent. This structured flow inherently facilitates seamless downstream concatenation while eliminating semantic cross-talk.

To further fortify the robustness of these compressed representations, we employ a \textit{Compression-Concatenation Co-training} strategy. By supervising the model on generation tasks conditioned on concatenated KV segments, we implicitly steer the \methodprefix Extractor to produce ``merge-ready'' states that maintain semantic coherence even when reordered. Consequently, \methodprefix KV enables aggressive compression with minimal fidelity loss, significantly alleviating the storage and bandwidth bottlenecks of long-context inference.

Our main contributions are summarized as follows:
\begin{itemize}[leftmargin=9pt,topsep=0pt]
\item We show that the key challenge of KV cache reuse lies in storage and memory transfer overheads, whereas existing approaches only reduce recomputation and ignore the storage overhead.
\item We propose a unified training formulation that jointly optimizes KV extraction and inference-time KV concatenation, explicitly aligning the extracted representations with their reuse behavior under non-prefix inference.
\item We introduce a new attention information flow through a carefully designed attention mask to preserve generation accuracy when concatenating compressed KV caches.
\item Extensive experiments show that our method significantly reduces KV cache storage and transfer cost, achieving up to 17$\times$ inference speedups over existing approaches, without degrading the inference quality.
\end{itemize}

%% file: text/2_background.tex
\section{Preliminaries and Motivation}\label{sec:2_background}
% \junyi{
% 2.1 LLM \& KV Cache \\
% 2.2 Opportunities of KV reuse\\
% 2.3 KV Reuse (Training-free methods: concat) \\
% 2.4 KV Reuse (Training-based methods: reuse compressed cache) \\
% 2.5 Motivation and Problem formulation \\
% }

% This section introduces the necessary background on Transformer architectures, prefix-based KV cache reuse, and non-prefix KV cache reuse. We then analyze the limitations of existing non-prefix KV cache reuse methods and formally characterize the underlying problem they face. 
% \shengzhong{Address the conclusion we want to make within each subsection, which have to be connected and lay the basis to understand the framework section.}

% \shengzhong{Evaluate what the necessary background, preliminaries, and motivation are to understand \model better. What should be the best order to organize them?}

\subsection{KV Cache in LLM Inference}

The Transformer-based Large Language Model (LLM) generates tokens auto-regressively. To avoid redundant computations, Key-Value (KV) caching stores the intermediate attention states. Formally, for a sequence $\{x_1, \dots, x_t\}$, the KV cache at layer $l$ is:

\begin{equation}
    \mathbf{K}_l = [k_1, \dots, k_t], \quad \mathbf{V}_l = [v_1, \dots, v_t].
\end{equation}

During the generation of $x_{t+1}$, the model only computes the KV pair for the new token and concatenates it with the existing cache. While accelerating standard inference, the sequential dependency of self-attention poses a challenge for reusing KV blocks in different contexts.

\subsection{Generalized KV Cache Reuse}

Existing LLM serving systems primarily utilize Prefix Caching (\eg vLLM~\cite{kwon2023efficient}, RadixAttention~\cite{NEURIPS2024_724be447}), which allows the reuse of KV states if the leading sequence of a new request exactly matches a previously cached one. This is achieved by simple position-aligned mapping. However, in complex scenarios such as multi-document RAG~\cite{kim2024autorag} or modular tool-use~\cite{lu2025toolsandbox}, reusable content (\eg $\text{Doc}_\text{A}, \text{Doc}_\text{B}$) may appear in arbitrary orders or be interleaved with dynamic instructions. To address this, we move beyond prefix matching and consider a Non-Prefix Caching paradigm.

We define a generalized two-stage framework for non-prefix reuse. Given a request with a static context $\mathbf{x}^{\text{s}}$, several independent cached documents $\{\mathbf{x}^{\text{d}}_1, \dots, \mathbf{x}^{\text{d}}_K\}$, and a query $\mathbf{x}^{\text{q}}$, the process is formalized as:

\begin{itemize}[leftmargin=9pt]
\item \textbf{Encoding}: An encoder $\mathcal{M}$ (typically the base LLM itself or a specialized module) extracts KV representations from documents: 
\begin{equation}
    \mathbf{KV}^{\text{reuse}}_k = \mathcal{M}(\mathbf{x}^{\text{d}}_k).
\end{equation}
\item \textbf{Blending}: A blender $\mathcal{B}$ adapts and integrates these pre-computed states into the new request context:
\begin{equation}
    \mathbf{KV}^{\text{total}} = \mathcal{B}(\mathbf{x}^{\text{s}}, \mathbf{KV}^{\text{reuse}}_1, \dots, \mathbf{KV}^{\text{reuse}}_K).
\end{equation}
\item \textbf{Inference}: The LLM computes the final output: 
\begin{equation}
    y = \mathcal{F}([\mathbf{KV}^{\text{total}}, \mathbf{x}^{\text{q}}]).
\end{equation}
\end{itemize}

Based on the implementation of $\mathcal{M}$ and $\mathcal{B}$, current methods fall into two categories:

\begin{itemize} [leftmargin=9pt]
\item \textbf{Training-Free Methods}: These methods (\eg CacheBlend~\cite{10.1145/3689031.3696098}, KVShare~\cite{yang2025kvshare}) use the vanilla LLM as the encoder and employ selective recomputation as the blender to reconcile mismatched KV states.

\item \textbf{Training-Based Methods}: These methods (\eg Block-Attention~\cite{ma2025blockattention}, KVLink~\cite{yang2025kvlink}) modify the model's architecture or fine-tune its weights so that it natively learns to process disjoint or reordered KV blocks.
\end{itemize}

\subsection{Limitations of Existing Reuse Methods}

Despite their potential, both categories of methods face critical bottlenecks that prevent efficient and lossless serving.

\begin{figure}
    \centering
    \includegraphics[width=\linewidth]{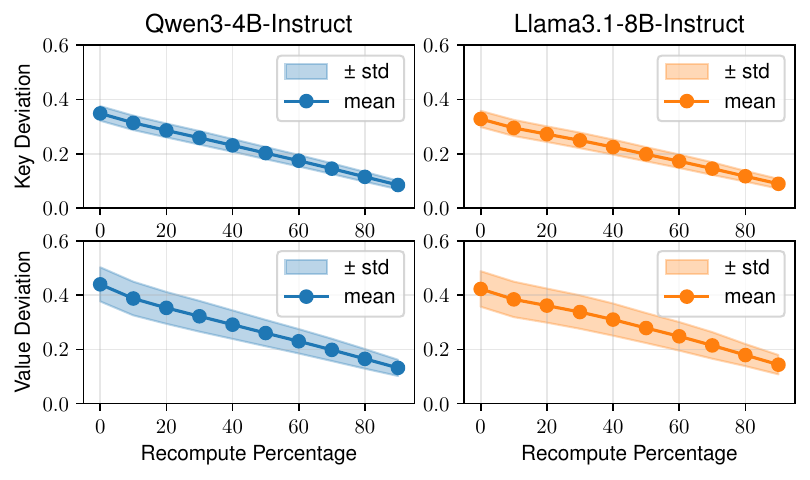}
    \caption{KV Deviation grows as recompute ratio drops.}
    \label{fig:kv_diff}
\end{figure}

\subsubsection{\textbf{The KV Deviation Gap in Training-Free Methods.}} 
In training-free methods, the blender $\mathcal{B}$ identifies a subset of tokens $M(\mathbf{x})$ to recompute. However, since the remaining unrecomputed KV states were generated in a different context, there exists an inherent KV deviation ($\Delta \mathbf{KV}$):
\begin{equation}
    \Delta \mathbf{KV} = \bigl\lVert \mathbf{KV}^{\text{full}} - \mathbf{KV}^{\text{blend}} \bigr\rVert\ /\ \bigl\lVert \mathbf{KV}^{\text{full}}\bigr\rVert \ge \delta > 0.
\end{equation}
As shown in Figure~\ref{fig:kv_diff}, this deviation monotonically increases as the recomputation ratio decreases, creating a rigid trade-off between serving latency and model accuracy.

\subsubsection{\textbf{Generalization Collapse in Training-Based Methods.}}
Training-based methods attempt to eliminate this gap by altering the self-attention mechanism, which suffers from:
\begin{itemize} [leftmargin=9pt]
\item \textbf{Capability Loss}: Altering attention patterns often leads to ``catastrophic forgetting''. Our empirical study shows a Llama3.1-8B model fine-tuned with Block-Attention suffers a performance drop of up to $10.4\%$ on LongBench.
\item \textbf{Inflexibility}: The high cost of per-model fine-tuning makes it difficult to adapt these methods to the rapid iteration of open-source base models.
\end{itemize}

\subsubsection{\textbf{Storage and Bandwidth Bottlenecks.}}

\begin{figure}
    \centering
    \includegraphics[width=0.95\linewidth]{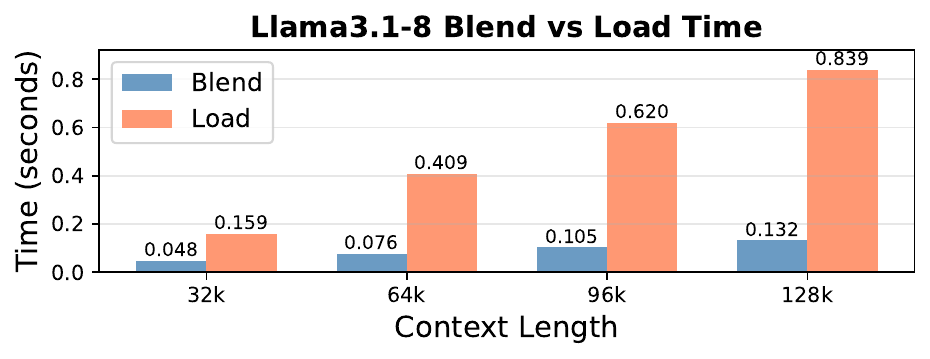}
    \caption{Lantecy comparison of blending and loading.}
    \label{fig:blend_vs_load}
\end{figure}

% \junyi{Figure~\ref{fig:prefill_vs_load} Some micro experiments need to be added to explain our motivation}
Crucially, prior methods primarily focus on reducing blending computation, while overlooking the fact that modern long-context serving is increasingly \textbf{memory-bound}. As context length $L$ grows, the KV cache size scales linearly, making the transfer of uncompressed KV states between memory hierarchies (\eg from DRAM to HBM) the dominant latency factor. As illustrated in Figure~\ref{fig:blend_vs_load}, existing reuse methods reduce prefill cost by loading cached KVs from memory, but still do not address this storage pressure; moreover, naively applying KV compression to reused blocks leads to severe fidelity loss, as standard compressed representations are not designed to be composable or position-agnostic.

% Figure~\ref{fig:ttft_breakdown} illustrates the TTFT breakdown under non-prefix document insertion. Existing non-prefix reuse methods reduce prefill cost by loading cached KVs from memory, but still incur substantial latency from transferring large, uncompressed KV states and performing additional blending operations, making TTFT memory- and bandwidth-bound. Moreover, Simply compressing reused KVs is insufficient, as standard compression is not designed for direct concatenation or position-agnostic reuse and leads to accuracy degradation. This motivates KV representations that are both compressible and directly composable.

%% file: text/3_framework.tex
\section{\model Framework} 
 \label{sec:3_framework}

\subsection{Framework Overview}

\begin{figure}[t!]
    \centering
    \includegraphics[width=0.95\linewidth]{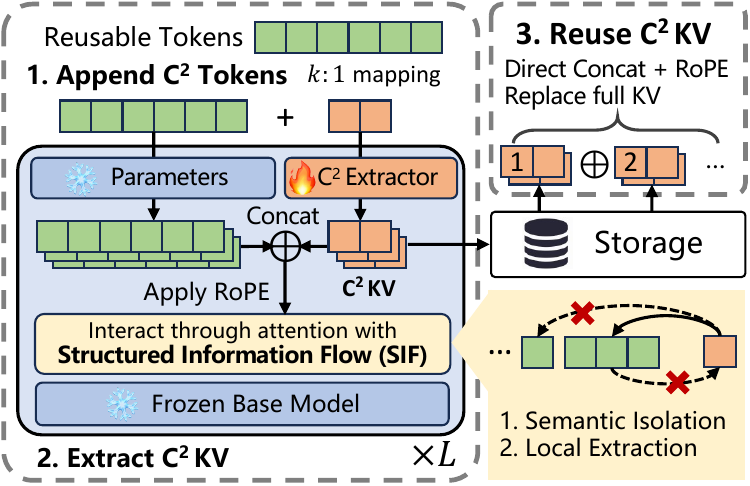}
    \caption{Pipeline overview of \method.}
    \label{fig:method_pipeline}
\end{figure}

We propose a non-prefix KV cache reuse framework \method that enables long documents to be compressed, stored, and directly reused at arbitrary positions. \method augments a frozen base LLM with a lightweight trainable module, together forming the \methodprefix Extractor, which produces compact and composable KV representations.

The pipeline consists of two stages. 
First, given a set of documents, the \methodprefix Extractor encodes each document independently into a compressed, position-agnostic KV cache, referred to as \methodprefix KV. 
The \methodprefix Extractor employs Structured Information Flow to preserve the base model’s semantics while enabling compression. Second, at inference time, the requested \methodprefix KV caches are retrieved, assigned positional embeddings according to their designated position, and directly concatenated to condition generation.

To ensure the validity of the extracted KV caches after concatenation, the \methodprefix Extractor is trained using Compression-Concatenation Co-Training, jointly optimizing compression and generation under a single supervised objective. 
This design decouples document encoding from query-time generation, enabling efficient non-prefix reuse with reduced KV cache size and minimal impact on generation quality.
An overview of the pipeline is illustrated in Figure~\ref{fig:method_pipeline}.

\subsection{\methodprefix Extractor}

\begin{figure*}[ht!]
    \centering
    \includegraphics[width=0.85\linewidth]{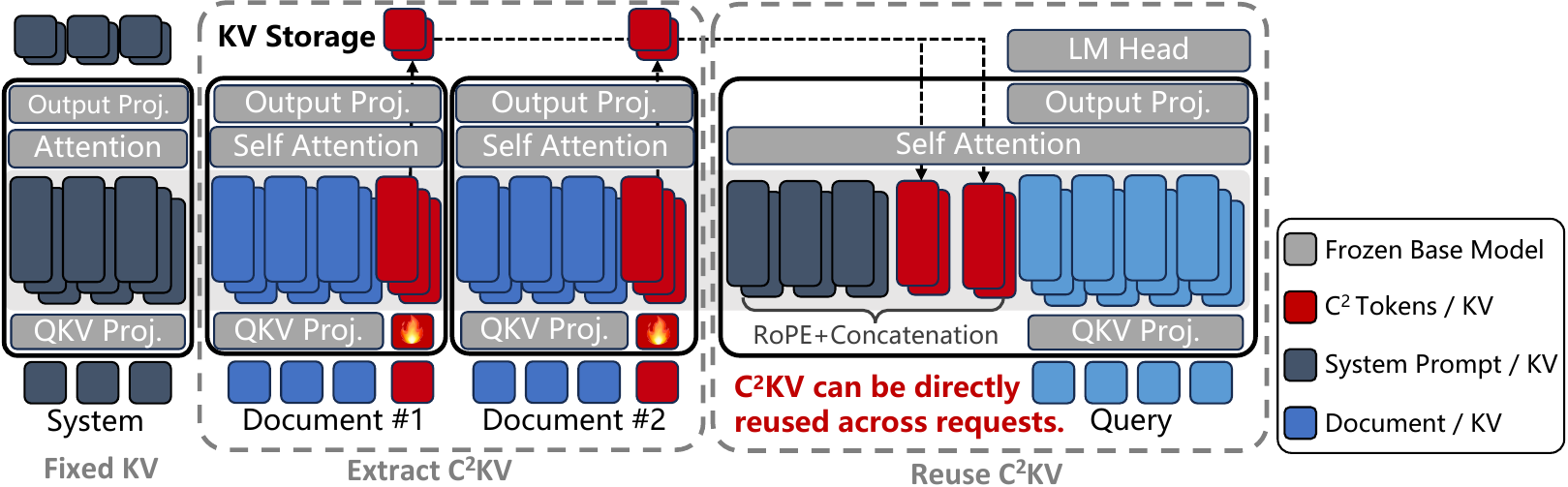}
    \caption{Architecture overview of \method.}
    \label{fig:method_architecture}
\end{figure*}

The \methodprefix Extractor is a lightweight module attached to a frozen base language model. 
It takes a document as input and produces a compressed, composable key-value representation, whose length is reduced by a $k\!:\!1$ ratio. 
Given a document token sequence, the \methodprefix Extractor outputs per-layer key-value pairs, referred to as \textbf{\methodprefix KVs}, which are stored and later retrieved for downstream generation.

\subsubsection{\textbf{\methodprefix Tokens as Memory Slots}} 
Compression is achieved by introducing a sequence of auxiliary tokens, referred to as \methodprefix Tokens, into the document token stream. Given a compression ratio k, the number of \methodprefix Tokens is determined as
m = $\lceil n / k \rceil$.
The \methodprefix Tokens are logically interleaved with the document tokens such that one \methodprefix Token is associated with each consecutive block of $k$ document tokens. 
All \methodprefix Tokens share the same learnable embedding vector and do not correspond to any lexical content.
% \shengzhong{Do we use a fixed embedding for all \methodprefix Tokens?}

% Each \methodprefix Token is assigned the same position index as the last token in its associated block (i.e., positions $k-1, 2k-1, \ldots$). This positional assignment is used only during extraction to define local alignment between document tokens and compression tokens. Importantly, the extracted KVs are stored before positional embeddings are applied, enabling the same KV cache to be reused later with different positional assignments.

\subsubsection{\textbf{New QKV Projection Heads}}

The core design of the \methodprefix Extractor is a set of new, per-layer QKV projection heads that operate exclusively on the \methodprefix Tokens. For each Transformer layer $\ell$, the base model’s original QKV projections are kept frozen and are applied only to the original document tokens. In contrast, \methodprefix Tokens are projected using separate trainable matrices:
\begin{equation}
Q^{(\ell)}_{\text{C}^2} = W^{(\ell)}_{{\text{C}^2},Q} h^{(\ell)}_{\text{C}^2}, \quad
K^{(\ell)}_{\text{C}^2} = W^{(\ell)}_{{\text{C}^2},K} h^{(\ell)}_{\text{C}^2}, \quad
V^{(\ell)}_{\text{C}^2} = W^{(\ell)}_{{\text{C}^2},V} h^{(\ell)}_{\text{C}^2},
\end{equation}
where the projection dimensions exactly match those of the base model, ensuring compatibility with downstream attention operations.

The projected \methodprefix QKV tensors are concatenated with the frozen QKV tensors of the original document tokens and passed into the standard attention operator. Only the parameters of the \methodprefix Token embeddings and their associated QKV projection heads are trainable; all base model parameters remain frozen. This design adds around $10\%$ of additional parameters to the base model for Qwen3-4B and Llama3.1-8B.

\subsubsection{\textbf{Residual Mean-Pooling Variant}} 
We also consider a variant of the Extractor that incorporates residual semantic aggregation, denoted as \methodprefix KV-Residual. In this variant, at the first Transformer layer, we compute the mean of the hidden states of the k document tokens associated with a given \methodprefix Token and add it to the \methodprefix Token’s hidden state:
\begin{equation}
h^{(\ell)}_{\text{C}^2} \leftarrow h^{(\ell)}_{\text{C}^2} + \frac{1}{k} \sum_{i \in \mathcal{B}} h^{(\ell)}_i,
\end{equation}
where $\mathcal{B}$ denotes the corresponding document token block. The scaling factor is fixed to $1.0$. This residual pathway provides an explicit aggregation of local document semantics and improves robustness under compression.

\subsubsection{\textbf{Output and Reuse}}
After extraction, only the KV pairs associated with the \methodprefix Tokens are retained as \methodprefix Extractor KVs. During inference, the retrieved KVs are applied new positional embeddings according to their placement in the final input sequence, where positional offsets are accumulated across concatenated segments and each compressed KV segment is treated as occupying the same effective span as its original uncompressed tokens. The compressed KV segments are then directly concatenated and used to condition downstream generation, without reintroducing the original tokens.

\subsection{Structured Information Flow}

The goal of Structured Information Flow (SIF) is to enable \textbf{semantic extraction and compression without altering the behavior of the frozen base model}. To this end, we impose a block-structured attention pattern that strictly separates the information pathways of Original Tokens and \methodprefix Tokens during extraction.

\subsubsection{\textbf{Token layout and block structure}}
Given a document token sequence $x_{1:n}$, we partition it into contiguous blocks of size $k$. For each block $\mathcal{B}(j)=\{(j-1)k+1,\ldots,\min(jk,n)\}$, we introduce exactly one \methodprefix Token $c_j$. Conceptually, $c_j$ is placed after all tokens in its corresponding block, so that standard causal attention suffices to enforce block-local visibility. For implementation convenience, all \methodprefix Tokens are physically grouped after the document tokens, and logical block boundaries are enforced solely through the attention mask. Each \methodprefix Token is assigned the same position index as the last token in its block. This positional assignment is used only during extraction; the resulting key-value pairs are stored before positional embeddings are applied.

\subsubsection{\textbf{Attention constraints}}
At every layer, we apply an attention mask that enforces the following constraints.

\begin{itemize} [leftmargin=0.35cm]
\item \textbf{Original-token invariance.} Original Tokens are allowed to attend only to \textbf{other Original Tokens} under the standard causal mask. Attention from Original Tokens to any \methodprefix Token is strictly disallowed:
\begin{equation}
M_{o,c}=-\infty \quad \forall\, o\in\mathcal{O},\; c\in\mathcal{C}.
\end{equation}
As a result, the hidden states and KV caches of Original Tokens are identical to those produced by the frozen base model.

\item \textbf{Block-local semantic extraction.}
Each \methodprefix Token $c_j$ may attend only to the Original Tokens in its own block $\mathcal{B}(j)$ and the sink block $\mathcal{B}(0)$:
\begin{equation}
M_{c_j,o}=0 \;\; \forall\, o\in\mathcal{B}(j)\cup\mathcal{B}(0),\ 
M_{c_j,o}=-\infty \;\; \forall\, o\notin\mathcal{B}(j)\cup\mathcal{B}(0).
\end{equation}
Since $c_j$ is conceptually positioned at the end of the block, causal masking guarantees that it can attend to all tokens within $\mathcal{B}(j)$ while preventing access to tokens beyond the block boundary. This enforces a $k\!:\!1$ structure and prevents global information leakage during extraction.

\item \textbf{Causal accumulation across blocks.} To allow compressed representations to accumulate document-level context, \methodprefix Tokens may attend to preceding \methodprefix Tokens causally. This enables later blocks to incorporate information from earlier compressed segments, while preserving autoregressive consistency.
\end{itemize}

\begin{figure}[t!]
    \centering
    \includegraphics[width=0.85\linewidth]{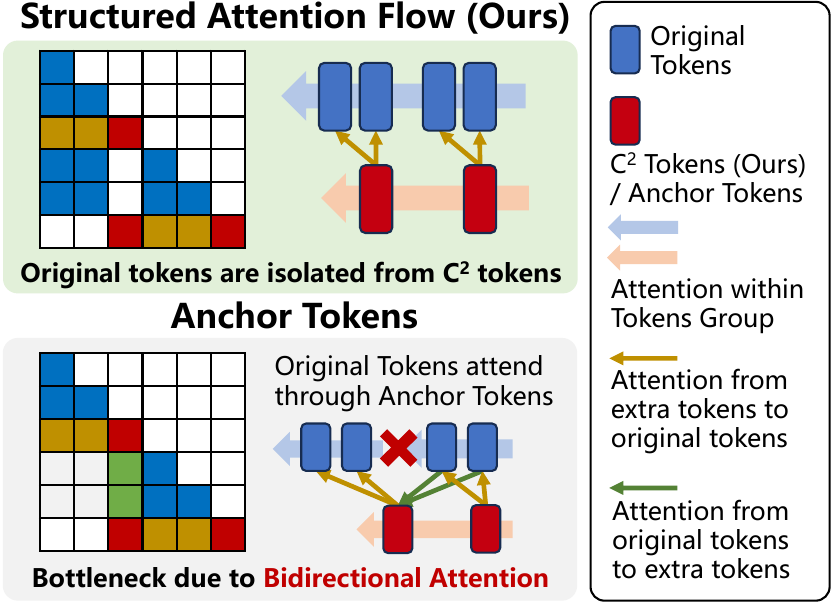}
    \caption{Comparison between Structured Attention Flow with \methodprefix Tokens and anchor token-based compression. The sink block in block-local extraction is omitted for simplicity.
    }
    \label{fig:anchor_tokens}
\end{figure}

\subsubsection{\textbf{Comparison with Anchor Token-based Compression.}}
Figure~\ref{fig:anchor_tokens} contrasts our Structured Attention Flow with anchor token-based compression methods, which also interleave auxiliary tokens into the input sequence. While both approaches introduce extra tokens, they serve fundamentally different purposes. Anchor tokens are designed as semantic bottlenecks for context compression and rely on bidirectional attention between original tokens and anchor tokens. As a result, original token representations are modified through anchor tokens, and the resulting KV cache is inherently entangled with the full input context.

In contrast, \methodprefix Tokens are introduced as latent KV carriers rather than semantic bottlenecks. Information flows only from original tokens to \methodprefix Tokens in a block-local and causal manner. This asymmetry is critical: it allows the extracted \methodprefix KVs to be context-independent, composable, and directly reusable across requests. Consequently, although anchor token-based methods are effective for compressing context within a single input, they are fundamentally incompatible with non-prefix KV cache reuse. Our design replaces semantic compression with structural KV extraction, enabling reuse by construction rather than approximation.

\subsection{Composable KV Learning via Compression-Concatenation Training}

The objective of compression-concatenation training is to ensure that document-level compression and KV cache concatenation are \textbf{}, so that independently extracted KV caches remain valid and useful after being composed in arbitrary multi-document contexts.

\subsubsection{\textbf{Training objective}} The entire model is trained using a single supervised fine-tuning objective on the generated answer. Let $\mathbf{K}_{\text{cat}}, \mathbf{V}_{\text{cat}}$ denote the concatenated document KV cache, and let $q_{1:m}$ and $y_{1:T}$ denote the query and answer tokens. 
The training loss is the standard autoregressive language modeling loss:
\begin{equation}
    \mathcal{L}_{\text{SFT}}
= - \sum_{t=1}^{T} \log p\bigl(y_t \mid y_{<t}, q_{1:m}, \mathbf{K}_{\text{cat}}, \mathbf{V}_{\text{cat}}\bigr).
\end{equation}
No auxiliary reconstruction loss, alignment loss, or document-level supervision is introduced. Gradients are propagated exclusively to the parameters of the \methodprefix Extractor, including the \methodprefix Token embeddings and their dedicated QKV projection heads; all base model parameters remain frozen.

\subsubsection{\textbf{Why concatenation is learned}} 
Although documents are extracted independently, supervision is applied only after concatenation. Each document’s \methodprefix KV cache is optimized based on its contribution to answering the query in the presence of other documents. Since document order and the number of documents vary across training samples, the Extractor is forced to project document semantics into a KV cache manifold that remains valid under different compositions.
Importantly, compression and concatenation are not optimized in isolation. A document KV cache that compresses too aggressively but fails after concatenation will incur higher generation loss, as will a KV cache that only functions in a fixed global context. Through end-to-end supervision, the Extractor learns a representation that balances both requirements and maps document tokens into a compression concatenation enabled KV manifold.

%% file: text/4_experiments.tex
\section{Evaluations} \label{sec:5_experiments}

\subsection{Experiment Setup}

\subsubsection{\textbf{Dataset}}
We train the \methodprefix Extractor using multi-document supervised fine-tuning (SFT) question-answering data, while keeping all Base Model parameters frozen.
The training corpus is constructed from the training splits of three publicly available QA datasets: HotpotQA~\cite{yang-etal-2018-hotpotqa}, 2WikiMultiHopQA~\cite{ho-etal-2020-constructing}, and LongMagpie~\cite{gao2025longmagpie}.
Together, these datasets cover a wide range of multi-document reasoning scenarios, making them suitable for learning document-level KV extraction under non-prefix reuse settings.

Each instance is formatted using the target model’s official chat template, including a system prompt, documents, user question, and reference answer. To improve formatting robustness and reduce verbosity, both system prompts and QA instructions are randomly sampled from a small prompt pool. During extraction, document tokens do not attend to system tokens; system KVs are concatenated only at generation time.

For evaluation, we use the LongBench~\cite{zhang-etal-2025-longcite} benchmarks.
For the single-document subsets in LongBench, we partition each document into natural paragraphs and treat them as multiple documents, enabling evaluation of non-prefix KV reuse.

% \noindent $\blacktriangleright$ \textbf{Training Dataset.} We train the Extractor using multi-document supervised fine-tuning (SFT) question-answering data, while keeping all Base Model parameters frozen. The training corpus is constructed from the training splits of three publicly available multi-document QA datasets: HotpotQA~\cite{yang-etal-2018-hotpotqa}, 2WikiMultiHopQA~\cite{ho-etal-2020-constructing}, and LongMagpie~\cite{gao2025longmagpie}). 
% From each dataset, we uniformly sample approximately 40,000 instances.
% Each training instance is reformatted into a unified structure consisting of a system prompt, a set of documents (provided by the dataset), a question, and a reference answer. We restrict the number of documents to at most 10 per instance, truncate each document to 1,024 tokens, and limit the combined length of the question and answer to 512 tokens (measured by the model tokenizer). System prompts and question prompts are augmented with variety.

% \noindent $\blacktriangleright$ \textbf{Evaluation Dataset.} We evaluate our method on the LongBench~\cite{zhang-etal-2025-longcite} benchmark and GSM8K~\cite{} benchmark. For the single-document subsets of LongBench, we cut the document into natural paragraphs and test them as multi-documents one. For GSM8K, we follow the strict evaluation setting and consider a prediction correct only if the final numeric answer exactly matches the reference, so as to test the model's generation ability.

\begin{figure*}[ht!]
    \centering
    \includegraphics[width=0.95\linewidth]{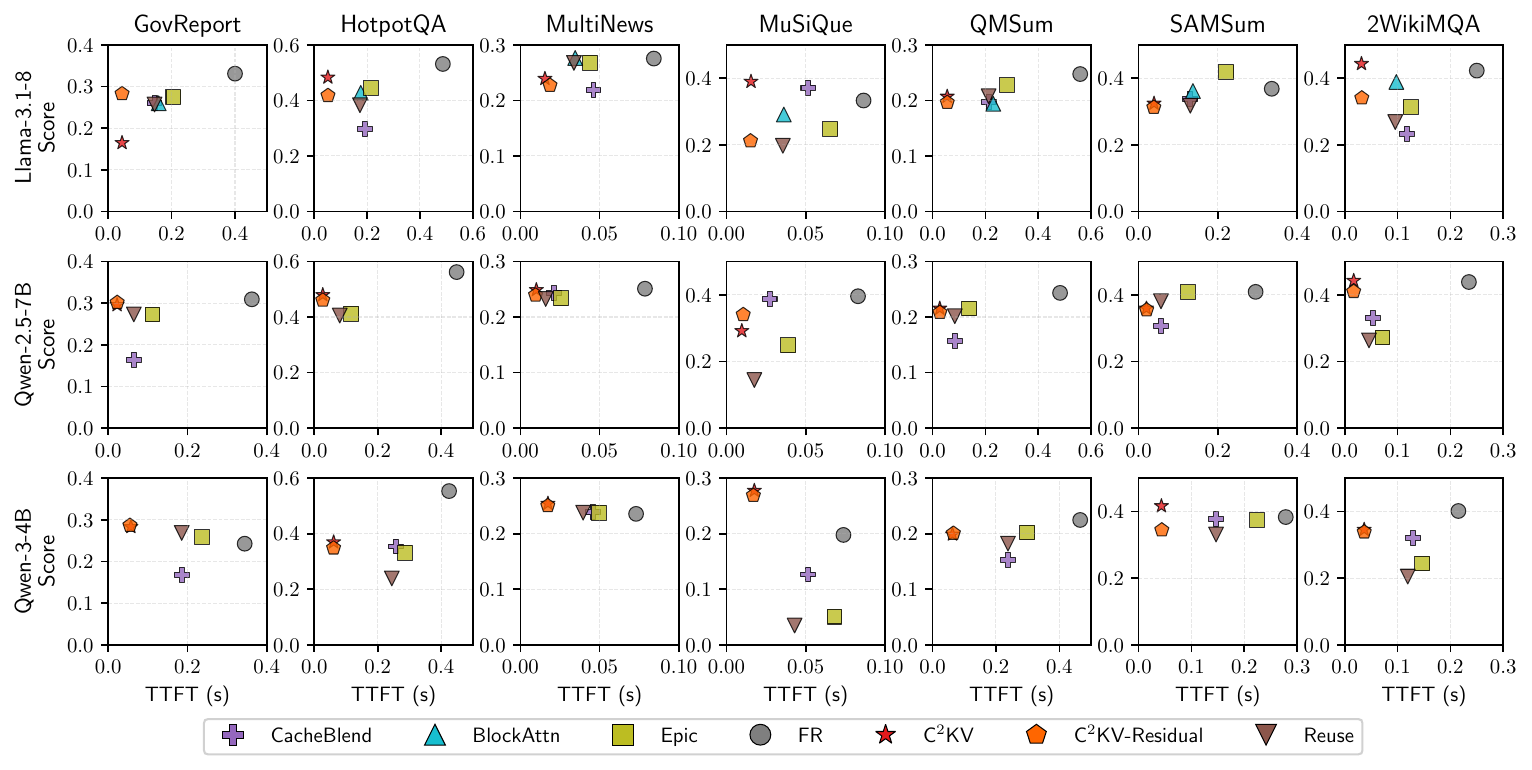}
    % \caption{Trade-off between Time-to-First-Token (TTFT) and task accuracy under non-prefix document reuse. We report dataset-level average TTFT (x-axis) and task-specific scores (y-axis) across seven LongBench tasks and three instruction-tuned LLMs.}
    \caption{Trade-off between Time-to-First-Token (TTFT, x-axis) and task accuracy (Score, y-axis) under non-prefix document reuse.}
    \label{fig:ttft_vs_score}
\end{figure*}

\subsubsection{\textbf{Metrics}} We adopt task-appropriate evaluation metrics following prior works.
For question-answering tasks, we report the \textbf{F1 score}, which measures token-level overlap between the predicted answers and the ground-truth references.
For summarization tasks, we report \textbf{ROUGE-L}, which evaluates sequence-level similarity based on the longest common subsequence between the generated and reference summaries.

\subsubsection{\textbf{Models}} We conduct experiments using three widely adopted instruction-tuned large language models: Qwen3-4B-Instruct-2507~\cite{yang2025qwen3}, Llama-3.1-8B-Instruct~\cite{grattafiori2024llama}, and Qwen2.5-7B-Instruct~\cite{yang2025qwen2}.
These models cover different architectures and parameter scales, allowing us to evaluate the robustness of our method across model families.

\subsubsection{\textbf{Baselines}} We compare \method against 5 baselines: 
\begin{itemize}[leftmargin=12pt]
    \item \textbf{Full KV Recompute (FR)}: The model performs a full prefill for every request without any KV reuse, serving as an accuracy and computation upper bound;
    \item \textbf{Full KV Reuse (Reuse)}: KV caches of reusable documents are directly reused without correction or retraining, representing an aggressive but error-prone lower bound;
    \item \textbf{EPIC}~\cite{hu2025epicefficientpositionindependentcaching}: A selective recompute reuse method that choose tokens by a static pattern;
    \item \textbf{CacheBlend}~\cite{10.1145/3689031.3696098}: A selective recompute reuse method that recomputes tokens by value differences to mitigate errors.
    \item \textbf{Block-Attention}~\cite{ma2025blockattention}: A training-based baseline by continue-training with no cross-attention between documents.
\end{itemize}

\subsection{TTFT-Accuracy Trade-off}

% \subsubsection{\textbf{Experiment setup.}}
% All results are reported as dataset-level averages over the full evaluation set. For each sample, document KVs are extracted offline using the corresponding method and stored in host memory; offline extraction time is not included in TTFT. 

% Document composition follows the original dataset specifications when available. HotpotQA and 2WikiMQA samples contain up to 10 documents, with document lengths potentially exceeding 1024 tokens. MuSiQue samples contain approximately 20 documents with around 512 tokens each. For datasets without predefined document boundaries, documents are segmented into chunks of approximately 512 tokens. The maximum generation length is capped at 64 tokens.

% All experiments are conducted on a single NVIDIA H200 GPU with batch size 1 and no request concurrency. Model parameters, activations, and KV caches use BF16 precision throughout. Document KVs are transferred from host memory to GPU; for reuse-based methods, TTFT includes both KV loading time and, when applicable, GPU blending time (measured separately and summed). For \method, TTFT includes KV loading as well as positional re-assignment and RoPE re-rotation, which incur negligible overhead due to element-wise operations.

\subsubsection{\textbf{Experiment setup.}} 
In this experiment, we focus on Time-to-First-Token (TTFT) under a document-reuse serving scenario, where a document corpus is available and document KV caches can be pre-extracted offline. TTFT is defined as the time elapsed from receiving a user request to the moment when the language model has its past KV cache fully prepared on GPU and is ready to start decoding the first output token.

Under this definition, different methods incur TTFT from different sources:
\begin{itemize}[leftmargin=12pt, topsep=0pt]
    \item Full Recompute (FR) must prefill the entire document context online and thus its TTFT corresponds to the full-context prefill latency.
    \item Non-prefix reuse methods load precomputed document KVs from host memory and additionally perform blending or structural transformations before decoding.
    \item \method loads compressed document KVs from host memory and directly reassigns positions without any blending, reducing TTFT to a load-only operation.
\end{itemize}
Although different reuse methods may employ different KV compression ratios, this difference reflects their inherent design goals. In particular, \method jointly performs KV compression (4$\times$) and reuse.

\subsubsection{\textbf{TTFT-Accuracy Trade-off}}

Figure~\ref{fig:ttft_vs_score} shows the trade-off between time-to-first-token (TTFT) and task accuracy across seven LongBench tasks and three instruction-tuned LLMs. 
Each point corresponds to the average TTFT and task score of a method on the full test set, where lower TTFT and higher accuracy are preferable. This analysis reveals how different KV reuse strategies balance startup latency against generation quality:

\mypara{Full recompute offers strong accuracy but poor startup latency}
While FR consistently achieves the highest accuracy by prefilling the entire context online, it suffers from prohibitive TTFT. This establishes a performance ceiling for accuracy but highlights the critical need for efficient KV reuse in long-context workloads.

\mypara{Existing non-prefix reuse reduces TTFT but introduces accuracy instability}
Prior non-prefix reuse approaches reduce TTFT (shifting leftward in the plot) but introduce substantial accuracy instability. Methods like EPIC and CacheBlend, which rely on KV blending, require complex preparation and exhibit sensitivity to architectural changes, leading to degraded performance in multi-document reasoning.

\mypara{\method achieves a superior TTFT-accuracy trade-off}
\method achieves minimal TTFT and competitive accuracy simultaneously, occupying the optimal upper-left region of the trade-off plot. Its learnable, composable KV representation transforms preparation into a load-only operation, eliminating errors from blending. While the \method-Residual variant improves stability on specific tasks, the vanilla \method delivers the best overall efficiency-performance balance.

Finally, we note that this experiment isolates TTFT, which dominates user-perceived latency for long-context requests.
Together with the decode-time results presented in the next section, these findings show that \method improves both request startup latency and steady-state generation efficiency.

\begin{figure*}[t!]
    \centering
    \includegraphics[width=0.85\linewidth]{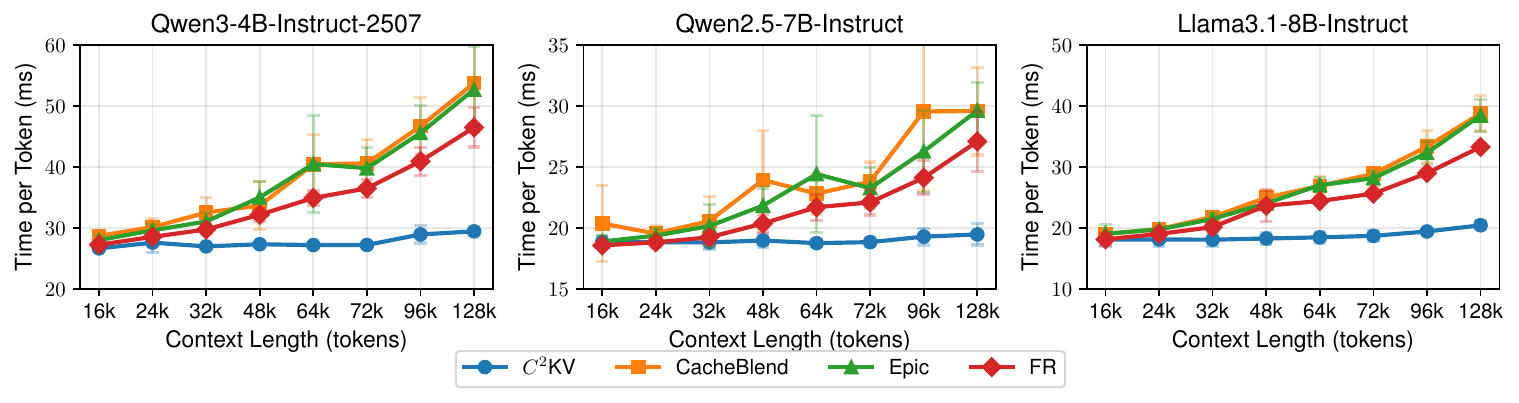}
    % \caption{Per-token decode latency under increasing context lengths. Offloading time grows nearly linearly with context length for full-length KV caches (FR), while \method significantly reduces transfer latency by shrinking the KV cache footprint, achieving a 2-3$\times$ reduction at long contexts.}
    \caption{Impact of context length on decode latency across different methods.}
    \label{fig:tbt-comparison}
\end{figure*}

\subsection{Decode-Time Efficiency}

% We evaluate system-level decode-time performance by measuring the average time per generated token (TBT), which directly reflects the memory bandwidth cost of KV cache access during autoregressive decoding.
% Prefill time is excluded from this measurement to isolate decode-stage behavior.
% All experiments are conducted with identical decoding settings, and KV cache reuse is applied before decoding starts.
% The results are shown in Figure~\ref{fig:tbt-comparison}.

We evaluate decode-stage efficiency using average time per token (TBT), isolating memory bandwidth costs by excluding prefill. All methods use identical decoding settings with KV reuse applied before decoding, and results are shown in Figure~\ref{fig:tbt-comparison}. Our findings are summarized as follows.

\mypara{\method significantly flattens decode-time scaling with context length}
As context length grows from 16k to 128k tokens, per-token decode latency increases for all methods due to longer attention contexts. Full-length KV caches exhibit near-linear scaling, indicating severe memory bandwidth pressure. In contrast, \method compresses the KV footprint and largely flattens the decode-time curve, showing only a mild increase even at 128k tokens across all models. Shrinking the effective KV cache size directly translates into lower time-between-tokens (TBT) under long-context decoding.
% As the context length increases from 16k to 128k tokens, all methods exhibit higher per-token decode latency due to the growing attention context.
% With full-length KV caches, per-token decode time increases nearly linearly with context length, reflecting escalating memory bandwidth pressure.
% In contrast, by compressing the KV cache footprint, \method substantially flattens the decode-time curve across all three models, with only a mild increase even at 128k-token contexts.

%\mypara{Decode-time gains are driven by reduced memory bandwidth consumption}
%Because prefill computation is excluded and KV reuse is applied prior to decoding, the observed improvements arise purely from reduced memory traffic during attention.
%This confirms that shrinking the effective KV cache size directly translates into lower time-between-tokens (TBT) under long-context decoding.

\mypara{KV compression complements prefill-stage reuse}
Beyond eliminating redundant prefill computation, \method delivers sustained performance benefits during decoding, where KV access dominates user-perceived latency.
These results highlight that non-prefix KV reuse, when combined with effective KV compression, yields system-level gains throughout the entire inference pipeline.

\subsection{Compression Ratio Scaling}

\input{table/compression_ratio_table}

We further evaluate how \method scales under different KV compression ratios on Llama3.1-8B. Table~\ref{tab:expr-compression-ratio} compares \method with existing non-prefix reuse methods under both fixed and dynamic compression settings.

%\mypara{Accuracy Comparable to Full Computation}
%Despite operating on compressed KV representations, \method maintains accuracy comparable to full-context computation across multiple tasks. Under \(4\times\) compression, \method achieves performance close to or exceeding the Full baseline on MuSiQue, WikiMQA, and SAMSum. These results suggest that the learned \(C^2\) representations preserve the key semantic information required for downstream generation.

\mypara{Scaling under Fixed Compression Ratios}
As the compression ratio increases from \(4\times\) to \(16\times\), \method exhibits graceful degradation and remains substantially more robust than existing reuse methods. Even under aggressive compression, \method preserves competitive performance across both QA and summarization tasks, demonstrating that the extracted KV representations remain stable and composable under large sequence reduction ratios.

\mypara{Generalization through Dynamic-Ratio Training}
We further train a single model by randomly sampling compression ratios from \(\{4\times, 8\times, 16\times\}\) at each training step. The resulting model generalizes effectively across different inference-time compression budgets, including the unseen \(10\times\) setting. Compared with fixed-ratio training, dynamic-ratio training consistently improves robustness and achieves stronger overall performance across varying compression levels.

Overall, these results demonstrate that \method provides a unified framework for composable KV reuse and scalable KV compression, while maintaining strong accuracy across a wide range of compression budgets.

\subsection{Long-Context Evaluation on RULER}

\begin{figure}[t!]
    \centering
    \includegraphics[width=\linewidth]{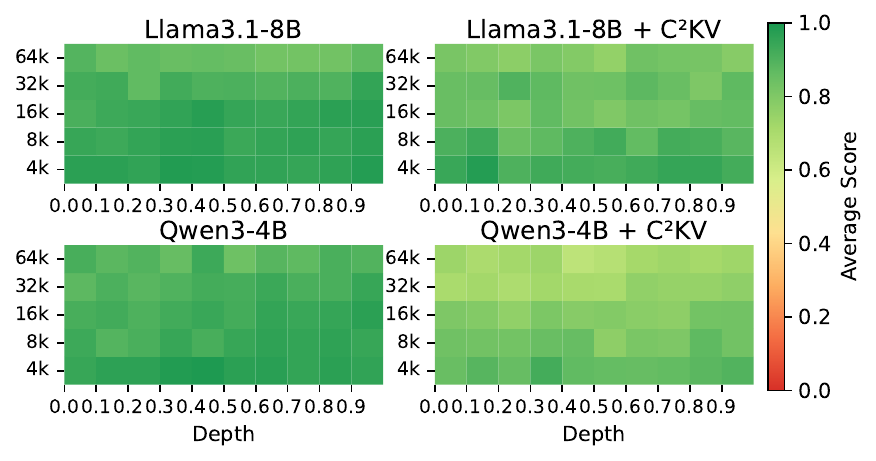}
    \caption{RULER evaluation across different context lengths and answer depths. \method is tested with 4$\times$ compression ratio. Depth denotes the relative position of the target answer span within the context, where larger depth values correspond to answers appearing later in the sequence.}
    \label{fig:ruler}
\end{figure}

We further evaluate \method on the RULER benchmark under 4\(\times\) KV compression. To support long-context evaluation, each input context is split into segments of approximately 4k tokens before extraction and KV reuse. Figure~\ref{fig:ruler} reports the average retrieval score across different context lengths and answer depths.

\mypara{Stable Long-Context Scaling}
\method maintains stable retrieval performance as the context length increases from 4k to 64k tokens, demonstrating effective long-range information preservation under compressed KV reuse.

%\mypara{Consistency across Answer Depths.}
%\method preserves retrieval behavior similar to the original full-context models across different answer depths, indicating that the compressed KV representations do not introduce additional positional bias.

\mypara{Robustness under Compression}
Despite operating with 4\(\times\) compressed KV caches, \method achieves only limited degradation compared with the original models, showing that composable KV compression can preserve strong long-context reasoning capability.

\subsection{Scaling to Larger Models}

\input{table/accuracy-qwen3-14b}

To evaluate whether \method scales to larger model sizes, we further conduct experiments on Qwen3-14B and report the results in Table~\ref{tab:accuracy-qwen3-14b}. We compare \method against existing non-prefix reuse baselines under multiple compression ratios.

The results show that \method consistently maintains strong performance across all tasks while enabling substantial KV compression. In particular, \(C^2\)KV-4x-Res achieves accuracy close to or exceeding the full-context baseline on MuSiQue, SAMSum, and MultiNews, while significantly outperforming existing reuse methods on multi-document QA tasks such as WikiMQA. Moreover, the performance degradation of \method remains gradual as the compression ratio increases from \(4\times\) to \(16\times\), demonstrating favorable scaling behavior under aggressive compression settings.

These results indicate that the composable KV representations learned by \method generalize effectively to larger LLMs and remain stable under high compression ratios.

\subsection{Does Compression Alone Suffice?}

\input{table/snapkv_degration}

To validate the claim that naively combining KV reuse with generic compression degrades accuracy, we conduct a motivating experiment. 
Table~\ref{tab:snapkv_degration} reports LongBench accuracy when a generic KV compression method, SnapKV ($4\times$ compression on all layers), is applied prior to KV reuse strategies.
Across all reuse methods, directly stacking SnapKV with KV reuse results in severe accuracy loss. 
This pattern is consistent across tasks and models, showing that generic KV compression, while reducing cache size, is incompatible with non-prefix KV reuse without composability-aware adaptation. 

These results motivate our design: compressing KV caches naively further distorts already non-composable KV representations, compounding errors during reuse. 
KV compression and reuse must therefore be jointly designed within a unified framework that produces KV representations both compressible and composable.

% \subsection{Information Retrieval Analysis}

% Multi Document Datasets have multi-hop question. Check if the models can retrieve information across all sub-questions.

\subsection{Ablation Study}

\input{table/ablation}

We conduct ablation studies to validate the key designs of \method. Table~\ref{tab:ablation} shows that replacing Structured Information Flow with bidirectional Anchor Tokens causes large accuracy degradation, indicating that unrestricted information mixing breaks KV composability. Directly using the untrained Extractor also fails severely, showing that composable KV representations must be learned.

\mypara{Position Re-Alignment}
\method (4x) Fixed-Pos. disables reuse-time positional re-rotation and consistently reduces accuracy, validating the necessity of position re-alignment after KV concatenation.

\mypara{Structural Constraints}
Relaxing block-local constraints through Info-Leakage or Global-Info attention harms performance across all tasks. These results show that stable KV reuse requires strict structural isolation and localized information aggregation.

Overall, the results support the core principles of \method: structured extraction, learned composable representations, and reuse-time positional alignment.

% Table~\ref{tab:ablation} shows that bidirectional anchor-token attention causes substantial accuracy drops, motivating Structured Information Flow that isolates original tokens from \methodprefix Tokens. Training is essential, as a naively initialized Extractor performs poorly. Reusing extraction-time positions without re-rotation degrades performance, supporting position-agnostic KV storage with reuse-time alignment. Finally, relaxing structured constraints—either allowing information flow from \methodprefix Tokens to original tokens or letting \methodprefix Tokens attend beyond local blocks—consistently reduces accuracy, confirming the importance of strict invariance and block-local extraction.

%% file: table/compression_ratio_table.tex
\begin{table}[t!]

\caption{Performance comparison on Llama3.1-8B across different KV cache methods and compression ratios. 
\small{C\(^2\)KV-\#x denotes models trained and evaluated with a fixed \#x KV compression ratio. 
C\(^2\)KV-Dyn-Test\#x denotes a single model trained with dynamically sampled compression ratios and evaluated under \#x inference-time compression ratio.}
}
\label{tab:expr-compression-ratio}

\resizebox{\linewidth}{!}{
\begin{tabular}{@{}l|cccc@{}}
\toprule
\multicolumn{1}{c|}{\textbf{LLama3.1-8B}} & \textbf{MuSiQue} & \textbf{WikiMQA} & \textbf{SAMSum} & \textbf{MultiNews} \\ \midrule
Full & 0.3198 & 0.4018 & 0.3652 & 0.2685 \\ \midrule
Naïve reuse & 0.1970 & 0.2681 & 0.3172 & 0.2674 \\
EPIC & 0.2498 & 0.2821 & 0.4217 & 0.2630 \\
CacheBlend & 0.2720 & 0.1334 & 0.3379 & 0.2126 \\ \midrule
\method-4x & 0.3587 & 0.4477 & 0.3904 & 0.2532 \\
\method-8x & 0.3225 & 0.4462 & 0.2572 & 0.2351 \\
\method-16x & 0.2746 & 0.3973 & 0.3556 & 0.2234 \\ \midrule
\method-Dyn-4xTest & 0.3457 & 0.4680 & 0.3779 & 0.2503 \\
\method-Dyn-8xTest & 0.3285 & 0.4555 & 0.3694 & 0.2428 \\
\method-Dyn-10xTest & 0.3198 & 0.4310 & 0.3570 & 0.2357 \\
\method-Dyn-16xTest & 0.2831 & 0.4298 & 0.3525 & 0.2343
\end{tabular}
}
\end{table}

%% file: table/accuracy-qwen3-14b.tex
% Please add the following required packages to your document preamble:
% \usepackage{booktabs}
\begin{table}[t!]
\caption{Performance comparison on Qwen3-14B across different non-prefix KV reuse methods and compression ratios. }
\label{tab:accuracy-qwen3-14b}
\resizebox{0.95\linewidth}{!}{
\begin{tabular}{@{}l|cccc@{}}
\toprule
\multicolumn{1}{c|}{\textbf{Qwen3-14B}} & \textbf{MuSiQue} & \textbf{WikiMQA} & \textbf{SAMSum} & \textbf{MultiNews} \\ \midrule
Full & 0.4384 & 0.5514 & 0.4060 & 0.2472 \\ \midrule
Naïve reuse & 0.1046 & 0.3184 & 0.2055 & 0.2421 \\
EPIC & 0.2645 & 0.3826 & 0.4093 & 0.2472 \\
CacheBlend & 0.4187 & 0.3955 & 0.3782 & 0.2413 \\ \midrule
\textbf{\method-4x} & 0.4311 & 0.5625 & 0.4105 & 0.2478 \\
\textbf{\method-8x} & 0.4043 & 0.5654 & 0.3835 & 0.2448 \\
\textbf{\method-16x} & 0.3342 & 0.5679 & 0.3871 & 0.2403 \\ \bottomrule
\end{tabular}
}
\end{table}

%% file: table/snapkv_degration.tex
% Please add the following required packages to your document preamble:
% \usepackage{booktabs}
\begin{table*}[h!]
\caption{Effect of naively combining KV cache compression (SnapKV) with reuse methods on Llama3.1-8B-Instruct.}
\resizebox{0.9\linewidth}{!}{
\begin{tabular}{@{}c|cc|cc|cc|cc|cc@{}}
\toprule
Dataset   & FR     & +SnapKV & Naive  & +SnapKV & BlockAttn & +SnapKV & Epic   & +SnapKV & CacheBlend & +SnapKV \\ \midrule
HotpotQA  & 0.5343 & 0.5045  & 0.3973 & 0.0547  & 0.4271         & 0.0743  & 0.4363 & 0.0897  & 0.2970     & 0.0532  \\
MuSiQue   & 0.3198 & 0.1615  & 0.1821 & 0.1366  & 0.2572         & 0.0657  & 0.2498 & 0.1799  & 0.2720     & 0.1311  \\
MultiNews & 0.2685 & 0.2421  & 0.2682 & 0.2179  & 0.2763         & 0.2202  & 0.2651 & 0.2300  & 0.2126     & 0.2177  \\
SAMSum    & 0.3652 & 0.3582  & 0.3151 & 0.1215  & 0.3599         & 0.1266  & 0.4217 & 0.1867  & 0.3379     & 0.1650   \\
2WikiMQA  & 0.4018 & 0.3438  & 0.2602 & 0.0936  & 0.3886         & 0.1396  & 0.2821 & 0.1681  & 0.1334     & 0.0765  \\ \bottomrule
\end{tabular}}

\label{tab:snapkv_degration}
\end{table*}

%% file: table/ablation.tex
% Please add the following required packages to your document preamble:
% \usepackage{booktabs}
\begin{table}[]
\caption{Ablation study of key design choices for composable KV cache extraction.}
\resizebox{0.9\linewidth}{!}{
\begin{tabular}{@{}lccc@{}}
\toprule
\multicolumn{1}{c}{Ablation Methods}                & Qasper & SAMSum & GovReport \\ \midrule
\multicolumn{1}{l|}{Base Model}                     & 0.4417 & 0.3652 & 0.3274    \\ \midrule
\multicolumn{1}{l|}{Anchor Tokens}                  & 0.2750 & 0.3201 & 0.1280    \\ \midrule
\multicolumn{1}{l|}{\method-Res. (4x) w/o T.} & 0.1572 & 0.3042 & 0.1318    \\
\multicolumn{1}{l|}{\method-Residual (4x)}          & 0.3185 & 0.3330 & 0.2802    \\ \midrule
\multicolumn{1}{l|}{\method (4x) Fixed-Pos.}       & 0.2827 & 0.3303 & 0.2801    \\
\multicolumn{1}{l|}{\method (4x) Info-Leakage}   & 0.2577 & 0.3406 & 0.2543    \\
\multicolumn{1}{l|}{\method (4x) Global-Info}   & 0.3722 & 0.3810 & 0.2888    \\
\multicolumn{1}{l|}{\method (4x)}                   & 0.3755 & 0.3904 & 0.2967    \\ \bottomrule
\end{tabular}}
\label{tab:ablation}
\end{table}

%% file: text/5_related.tex
\section{Related Work} \label{sec:9_related}
\mypara{KV Cache Reuse}Efficient KV cache reuse is critical for reducing Time-to-First-Token (TTFT) and computational redundancy. \textit{Prefix-based methods}, such as PromptCache~\cite{gim2024prompt}, PagedAttention~\cite{kwon2023efficient} and Radix-Attention~\cite{NEURIPS2024_724be447}, cache KV states of common prefixes in a tree structure to avoid recomputation across multiple requests. To handle non-prefix scenarios, \textit{training-free approaches} like CacheBlend~\cite{10.1145/3689031.3696098} and Cache-Craft~\cite{10.1145/3725273} enable arbitrary text chunk reuse via selective recomputation and KV state alignment, though they may suffer from slight accuracy degradation. \textit{Training-based methods}, including TurboRAG~\cite{lu-etal-2025-turborag} and Block-Attention~\cite{ma2025blockattention}, modify model architectures or employ fine-tuning to support modular cache stitching (\eg independent attention blocks). While these offer better generation quality, they require significant training overhead and risk compromising the base model's general capabilities.

\mypara{KV Cache Compression}
%To mitigate the memory bottleneck of Large Language Models (LLMs), KV cache compression techniques focus on reducing the storage footprint of each token. Quantization methods~\cite{hooper2024kvquant,zhang2024kv,tan2024alignedkv} reduce the bit-precision of KV tensors, often utilizing outlier-aware scaling to maintain performance. Sparsification or Eviction strategies, such as H$_2$O~\cite{zhang2023ho},  StreamingLLM~\cite{xiao2024efficient} and SnapKV~\cite{10.5555/3737916.3738638}, identify and retain only "heavy-hitter" or "anchor" tokens, discarding less significant ones based on attention scores. More recently, Token Merging approaches~\cite{10.5555/3692070.3694498} consolidate redundant KV states into a smaller set of representative vectors. Unlike reuse-centric methods, compression focuses on memory efficiency within a single context but can be combined with reuse to further scale long-context processing.
To mitigate LLM memory bottlenecks, KV cache compression reduces per-token storage. Quantization methods like CommVQ~\cite{li2025commvq} or MiniKV~\cite{sharma2025minikv} achieve ultra-low bit-precision (\eg 2-bit) while preserving accuracy. Eviction strategies, such as SnapKV~\cite{10.5555/3737916.3738638} and RocketKV~\cite{behnam2025rocketkv}, identify and retain crucial tokens via attention-driven scores or reconstruction importance. Furthermore, Merging and Low-rank approaches like MiniCache~\cite{liu2024minicache} or ReCalKV~\cite{yan2025recalkv} consolidate redundant KV states across layers or attention heads. These techniques maximize memory efficiency and, when paired with reuse, significantly scale long-context processing.

%% file: text/6_conclusion.tex
\section{Conclusion} \label{sec:10_conclusion}
In this work, we identify KV cache storage and memory bandwidth as the key bottlenecks in non-prefix KV reuse for long-context LLM inference. We show that existing reuse and compression methods are fundamentally limited by the non-composability of standard KV representations. To address this, we propose \method, a unified framework that learns a composable and compressed KV cache manifold through a lightweight extractor and structured attention, while keeping the base model frozen. Extensive experiments demonstrate that \method enables efficient KV reuse and compression without degrading generation quality, substantially reducing inference overhead under long contexts.
% We propose \model, a new form of working latent memory for LLM inference.
% Unlike KV cache reuse or architectural modifications, it introduces a reusable, composable memory abstraction while preserving base model behavior.

%% file: appendix/appendix_outline.tex
\section{Appendix Outline}
In this appendix, we report the following sections as supplementary materials for the main paper.

\begin{itemize} [leftmargin=0.55cm,topsep=0pt]
    \item Appendix~\ref{appendix:setup} shows the dataset setup for training and evaluation.
     % \item Appendix~\ref{appendix:implementation} introduces the metrics adopted in our evaluation results.
    \item Appendix~\ref{appendix:ttft} shows the detailed TTFT measurement protocol.
    \item Appendix~\ref{appendix:results} gives full results for accuracy evalution.
    \item Appendix~\ref{appendix:limitations} lists the limitations and future work.
\end{itemize}

%% file: appendix/setup.tex
\section{Dataset Setup} \label{appendix:setup}

\subsection{Training Setup}

We train the \methodprefix Extractor using supervised fine-tuning on multi-document instruction-following data sampled from HotpotQA, 2WikiMQA, and LongMagpie. We sample 40k instances from each dataset, resulting in 120k total training samples, and train for one epoch only. For LongMagpie, long passages are segmented into multiple documents using sentence- or paragraph-aware splitting. 

The base LLM remains frozen throughout training. Only the C$^2$ token embeddings and per-layer QKV projection heads are trainable. We use AdamW with cosine learning rate scheduling under the following settings: Learning rate: 5e-5, Warmup ratio: 0.06, Weight decay: 0.1, Global batch size: 32, Gradient accumulation: 4 and Precision: BF16.

\subsection{Evaluation Setup}

We evaluate \method on LongBench datasets covering multi-document QA and long-form summarization, including HotpotQA, 2WikiMQA, MuSiQue, MultiNews, SAMSum, QMSum, and GovReport.

For QA tasks, we report F1 following official evaluation protocols. For summarization tasks, we report ROUGE-L. All methods use greedy decoding, with maximum generation lengths of 32 tokens for QA and 512 tokens for summarization.

%% file: appendix/ttft_measurement.tex
\section{Detailed TTFT Measurement Protocol} \label{appendix:ttft}

This section formalizes the protocol used to measure Time-to-First-Token (TTFT) in our system-level evaluations.

\subsection{Definition and Timing Boundary}

TTFT is defined as the elapsed wall-clock time from when a user request arrives to when the language model has its past key–value cache fully prepared on GPU and is ready to start decoding the first output token.

We explicitly exclude the prefill cost of system prompts and query tokens from TTFT for all methods, as their lengths are short and negligible relative to long-context document processing. Offline document KV extraction time is also excluded. All timing measurements are synchronized using torch.cuda.synchronize() to ensure accurate GPU timing.

\subsection{Method-Specific TTFT Components}

Under this definition, TTFT is measured as follows:
\begin{itemize}[leftmargin=12pt, topsep=0pt]
    \item Full Recompute (FR): TTFT equals the wall-clock time of performing a full prefill over the entire document context to construct the KV cache, since non-prefix document reuse is not supported.
    \item Load+Blend Methods: TTFT is the sum of loading precomputed document KVs from host memory to GPU and executing GPU-side KV blending or structural transformation kernels. These components are profiled separately and summed.
    \item \model: TTFT includes loading compressed document KVs from host memory to GPU and lightweight positional reassignment and RoPE re-rotation. No blending or recomputation is required.
\end{itemize}

%% file: appendix/full_result.tex
\section{Full Results} \label{appendix:results}

\subsection{Accuracy Evaluation}

\input{table/accuracy}

We evaluate non-prefix KV cache reuse across multiple models and tasks under a unified setup. KV reuse is applied only to document tokens, while query and generation tokens are always computed online. Block-Attention results are reported only for Llama-3.1-8B-Instruct due to checkpoint availability.

As shown in Table~\ref{tab:accuracy}, \method maintains strong accuracy under $4\times$ KV compression, consistently approaching Full KV Recompute (FR) across models and tasks. This demonstrates that non-prefix KV reuse can remain accurate when using composable KV representations.

In contrast, directly concatenating standard KV caches causes severe degradation on multi-document QA tasks, revealing the non-composability of vanilla self-attention KV representations. Training-free reuse methods such as EPIC and CacheBlend partially improve over naive reuse but exhibit unstable performance across models and datasets.

Under the same compression ratio, \method achieves the best or near-best accuracy among reuse-based approaches on most tasks. Compared with generic KV compression methods such as SnapKV, \method attains comparable or better accuracy while additionally supporting composable KV reuse.

%% file: table/accuracy.tex
\begin{table*}[t!]
\centering
% \caption{Accuracy comparison on LongBench tasks and GSM8K across three instruction-tuned LLMs. Our method (\methodprefix KV) consistently achieves the best or near-best accuracy among reuse-based methods, demonstrating that non-prefix KV cache reuse can be enabled without degrading generation quality.
% \textit{Bold} indicates the best reuse-based result; \underline{underline} indicates the second best. The 1$\times$/4$\times$ markers indicate the KV cache compression ratio of these methods.}
\caption{Accuracy comparison on LongBench and GSM8K. Bold and \underline{underline} denote the best and second-best reuse results, respectively; $1\times/4\times$ indicate KV compression ratios.}
\label{tab:accuracy}
\resizebox{0.95\linewidth}{!}{
\begin{tabular}{l|c|c|c|c|c|c|c|c|c}
\toprule
 & \multicolumn{7}{c|}{\textbf{Multi-Document Information Retrieval}} & \multicolumn{2}{c}{\textbf{Few-shot Learning}} \\ \cmidrule(lr){2-8} \cmidrule(l){9-10}
 & \multicolumn{5}{c|}{\textbf{QA}} & \multicolumn{2}{c|}{\textbf{Summary}} & \textbf{QA} & \textbf{Summary} \\ \cmidrule(lr){2-6} \cmidrule(lr){7-8} \cmidrule(l){9-10}
 & \small HotpotQA & \small 2WikiMQA & \small MuSiQue & \small QMSum & \small Qasper & \small MultiNews & \small GovReport & \small GSM8K & \small SAMSum \\ \midrule

% --- Qwen3 Section ---
\rowcolor{Blue!10}
\multicolumn{10}{c}{\textbf{Qwen3-4B-Instruct-2507}} \\ \midrule
FR (1$\times$) & 0.5392 & 0.4095 & 0.1991 & 0.2419 & 0.3074 & 0.2386 & 0.2826 & 0.8953 & 0.3846 \\
FR-SnapKV (4$\times$) & 0.3715 & 0.2930 & 0.0821 & 0.2022 & 0.2055 & 0.2087 & 0.2602 & 0.9007 & 0.3697 \\ \midrule
Naive (1$\times$) & 0.2486 & 0.2035 & 0.0339 & 0.1820 & 0.2521 & 0.2388 & 0.2666 & 0.7448 & 0.3296 \\
Epic (1$\times$) & 0.3266 & 0.2412 & 0.0492 & {\ul 0.2151} & {\ul 0.2697} & 0.2364 & 0.2565 & \textbf{0.8999} & 0.3795 \\
CacheBlend (1$\times$) & {\ul0.3487} & 0.2522 & 0.2049 & 0.1573 & 0.1730 & 0.2398 & 0.1680 & \textbf{0.8999} & 0.3772 \\
\rowcolor{Teal!10} 
$C^2$KV-Res. (4$\times$) & 0.3482 & \textbf{0.4419} & {\ul0.2194} & \textbf{0.2175} & 0.2662 & {\ul 0.2506} & \textbf{0.2830} & 0.8837 & \textbf{0.3883} \\
\rowcolor{Teal!10} 
$C^2$KV (4$\times$) & \textbf{0.3801} & {\ul0.3920} & \textbf{0.2300} & 0.1824 & \textbf{0.3344} & \textbf{0.2529} & {\ul 0.2744} & 0.8730 & {\ul 0.3836} \\ \midrule

% --- Llama3 Section ---
\rowcolor{Blue!10}
\multicolumn{10}{c}{\textbf{Llama3.1-8B-Instruct}} \\ \midrule
FR (1$\times$) & 0.5343 & 0.4018 & 0.3198 & 0.2452 & 0.4417 & 0.2685 & 0.3274 & 0.6914 & 0.3652 \\
FR-SnapKV (4$\times$) & 0.5045 & 0.3438 & 0.1615 & 0.2201 & 0.2836 & 0.2421 & 0.2811 & 0.6990 & 0.3582 \\ \midrule
Naive (1$\times$) & 0.3973 & 0.2602 & 0.1821 & 0.2075 & {\ul 0.3804} & {\ul 0.2682} & 0.2573 & 0.4263 & 0.3151 \\
Epic (1$\times$) &{\ul 0.4363} & 0.2821 & 0.2498 & {\ul 0.2231} & \textbf{0.4109} & 0.2630 & 0.2782 & \textbf{0.7319} & \textbf{0.4217} \\
CacheBlend (1$\times$) & 0.2970 & 0.1334 &  0.2720 & 0.1967 & 0.2107 & 0.2126 & 0.1860 & 0.6895 & 0.3379 \\
Block-Attn (1$\times$) & 0.4271 & 0.3886 & 0.2572 & 0.2161 & 0.3433 & \textbf{0.2763} & 0.2595 & 0.6371 & 0.3599 \\
\rowcolor{Teal!10} 
$C^2$KV-Res. (4$\times$) & 0.4173 & \textbf{0.4526} & {\ul 0.3395} & \textbf{0.2310} & 0.3185 & 0.2448 &{\ul 0.2802} & {\ul 0.7189} & 0.3330 \\
\rowcolor{Teal!10} 
$C^2$KV (4$\times$) & \textbf{0.4828} & {\ul 0.4477} & \textbf{0.3587} & 0.1885 & 0.3755 & 0.2532 & \textbf{0.2967} & 0.6175 & {\ul 0.3904} \\ \midrule

% --- Qwen2.5 Section ---
\rowcolor{Blue!10}
\multicolumn{10}{c}{\textbf{Qwen2.5-7B-Instruct}} \\ \midrule
FR (1$\times$) & 0.5848 & 0.4332 & 0.3448 & 0.2266 & 0.4358 & 0.2475 & 0.3073 & 0.7991 & 0.4048 \\
FR-SnapKV (4$\times$) & 0.4871 & 0.3060 & 0.1829 & 0.1988 & 0.2755 & 0.2166 & 0.2858 & 0.8701 & 0.4027 \\ \midrule
Naive (1$\times$) & 0.3901 & 0.2700 & 0.1321 & 0.1820 & 0.3232 & 0.2314 & 0.2716 & 0.5737 & 0.3621 \\
Epic (1$\times$) & 0.4360 & 0.2610 & 0.2211 & {\ul 0.1992} & \textbf{0.3663} & 0.2361 & 0.2727 & 0.7212 & \textbf{0.4113} \\
CacheBlend (1$\times$) & 0.1903 & 0.3244 & 0.2224 & 0.1530 & 0.2001 & {\ul 0.2463} & 0.1642 & \textbf{0.8365} & 0.3063 \\
\rowcolor{Teal!10} 
$C^2$KV-Res. (4$\times$) & {\ul 0.4606} & {\ul 0.4655} & \textbf{0.3655} & \textbf{0.2152} & 0.3019 & \textbf{0.2486} & {\ul 0.2812} & {\ul 0.7502} & {\ul 0.3805} \\
\rowcolor{Teal!10} 
$C^2$KV (4$\times$) & \textbf{0.4795} & \textbf{0.4718} & {\ul 0.2856} & 0.1879 & {\ul 0.3491} & 0.2453 & \textbf{0.2987} & 0.7141 & 0.3754 \\\bottomrule
\end{tabular}}
\label{tab:accuracy}
\end{table*}

%% file: appendix/limitations.tex
\section{Limitations and Future Work} \label{appendix:limitations}

While \model effectively enables compressed and composable non-prefix KV reuse, it has several limitations that suggest directions for future work. First, \model focuses on document-level reuse and assumes that reusable content can be identified and extracted offline. Extending the extractor to support online or incremental KV extraction is an important direction. Second, our current design applies a uniform compression ratio across documents and layers. Future work could explore adaptive or content-aware compression strategies that preserve the composability property while allocating compression budgets more flexibly.